\documentclass[journal]{IEEEtran}
\usepackage{graphicx}
\usepackage{amsmath,amssymb,amsfonts}%
\usepackage{booktabs}
\usepackage[dvipsnames]{xcolor}
\usepackage[numbers]{natbib}
\usepackage{orcidlink}
\usepackage{balance}

\newcommand{\code}[1]{\textcolor{blue!70!black}{\texttt{#1}}}

\title{Conversational AI for Rapid Scientific Prototyping: A Case Study on ESA’s ELOPE Competition}

\author{Nils Einecke \orcidlink{0000-0002-2460-4955}
\thanks{Nils Einecke is with the Honda Research Institute Europe GmbH, Offenbach, Germany (e-mail: nils.einecke@honda-ri.de)}
}
\date{September 2025}

\begin{document}

\maketitle

\begin{abstract}
  Large language models (LLMs) are increasingly used as coding partners, yet their role in accelerating scientific discovery remains underexplored. This paper presents a case study of using ChatGPT for rapid prototyping in ESA's ELOPE (Event-based Lunar OPtical flow Egomotion estimation) competition. The competition required participants to process event camera data to estimate lunar lander trajectories. Despite joining late, we achieved second place with a score of 0.01282, highlighting the potential of human–AI collaboration in competitive scientific settings. ChatGPT contributed not only executable code but also algorithmic reasoning, data handling routines, and methodological suggestions, such as using fixed number of events instead of fixed time spans for windowing. At the same time, we observed limitations: the model often introduced unnecessary structural changes, gets confused by intermediate discussions about alternative ideas, occasionally produced critical errors and forgets important aspects in longer scientific discussions. By analyzing these strengths and shortcomings, we show how conversational AI can both accelerate development and support conceptual insight in scientific research. We argue that structured integration of LLMs into the scientific workflow can enhance rapid prototyping by proposing best practices for AI-assisted scientific work.
\end{abstract}

\section{Introduction}

The pace of scientific discovery is increasingly shaped by the ability to rapidly prototype algorithms, process large datasets, and test competing approaches. Modern scientific problems often demand solutions that combine insights from multiple disciplines and often reside in the digital domain, making software development a critical component of the research process. Yet coding for science is rarely straightforward: it requires not only implementing known algorithms, but also adapting them to novel data modalities, integrating diverse tools, and iterating quickly in the face of uncertainty. For many researchers, the bottleneck is not a lack of ideas, but the time and effort required to transform those ideas into working software.

Large language models (LLMs) such as ChatGPT offer a new way to overcome this bottleneck. Beyond generating executable code, they can explain underlying algorithms and theories, suggest alternative strategies, and assist with routine tasks like data conversion and screening of related-work. Their interactive nature enables a style of human–AI collaboration where conceptual discussion and practical implementation proceed hand in hand. In principle, such tools could accelerate the transition from scientific insight to working prototype, freeing researchers to focus more on creating ideas and result interpretation.

Despite this promise, the role of conversational AI in scientific working and prototyping remains underexplored. Most prior discussions of LLMs emphasize their use in software engineering, education, or general-purpose coding. Far less is known about how they perform in high-stakes scientific contexts, where accuracy, robustness, and domain-specific reasoning are critical. Understanding both the benefits and limitations of this new mode of collaboration is essential if these tools are to become a reliable part of a new scientific workflow.

To explore this question, we present a case study using ChatGPT as teammate in the ELOPE (Event-based Lunar OPtical flow Egomotion estimation) competition~\cite{Fanti_2026, esa_elope} organized by ESA's Advanced Concepts Team (ACT) in partnership with University of Adelaide and TU Delft. The competition required participants to process event camera data in order to estimate lunar lander trajectories. Entering late in the challenge, we faced a compressed development timeline — an ideal test for evaluating whether conversational AI could accelerate the scientific prototyping process. By pairing with ChatGPT, we achieved second place with a final score of 0.01282, demonstrating both the promise and the practical realities of AI-assisted coding in a competitive scientific setting.

Our experience highlights clear benefits: ChatGPT contributed functional code, algorithmic reasoning, and useful methodological suggestions, such as using fixed number of events instead of fixed time spans for windowing. At the same time, it revealed important limitations. The model frequently made unnecessary structural changes to code, got confused by intermediate discussions about alternative ideas, occasionally produced critical silent errors and forgot important aspects of the competition after some time of scientific discussion. These issues complicated development and demonstrate the need for careful human oversight and control of the scientific AI-assisted workflow.

By analyzing these strengths and shortcomings, we offer a balanced perspective on the potential of conversational AI for scientific prototyping. Our findings suggest that LLMs can meaningfully accelerate development while supporting conceptual exploration, but they also reveal current obstacles to smooth integration into a research workflow. We conclude by proposing recommendations for future AI-assisted research that take the current limitations of conversational AI into account.

\section{Related Work}
\label{sec:related}

Large Language Models (LLMs) have widely been studied as assistants for software engineering and coding-related tasks. Early work examined their role in code comprehension and analysis, where LLMs can support understanding complex codebases and assist in deductive coding tasks by generating explanations, documentation and categorizing content at scale \cite{Chew_2023, Nam_2024}. Newer investigations also analyzed the potential of LLMs in finding code smells \cite{Sadik_2025}. In educational contexts, research highlights that novices often rely on LLM-based generators to solve introductory programming problems, with mixed outcomes: while these tools provide scaffolding and accelerate problem-solving, beginners frequently struggle to interpret or adapt the generated code to their own understanding \cite{Kazemitabaar_2024, Zi_2025}. Beyond education, studies have investigated LLMs’ potential in professional and real-world coding environments. For instance, LLMs have been evaluated as freelance developers, showing promise in completing software tasks with economic value, albeit with limitations in robustness and reliability \cite{Miserendino_2025}. Similarly, analyses of student-LLM interactions in software engineering projects reveal that LLMs can augment productivity and collaboration, but also raise questions about over-reliance, code quality, and learning outcomes \cite{Naman_2025}. Complementing these qualitative insights, controlled experiments with GitHub Copilot provide quantitative evidence: developers with access to the tool completed a programming task about 55\% faster than those without, with particularly large gains for less experienced programmers \cite{Peng_2023}. Taken together, these findings suggest that LLMs hold substantial potential as coding assistants, yet their integration requires careful consideration of user expertise, interpretability, and context of use.

Meanwhile LLMs are also increasingly explored as tools for supporting scientific research. Surveys and prototypes of LLM-based agents show their ability to contribute across the research workflow, from hypothesis generation and experiment design to data analysis and manuscript drafting \cite{Joublin_2023, Ren_2025, Schmidgall_2025, Ifargan_2025}. These systems demonstrate efficiency gains and the potential for partial automation, but they also highlight challenges of reproducibility, verification, and ensuring genuine scientific novelty. Complementing these perspectives, empirical studies provide insight into how researchers already use LLMs in practice. Nejjar et al. report adoption for code generation, data cleaning, analysis, and visualization scripting, emphasizing both their practical value and recurring concerns about correctness, integrity, and variability of outputs \cite{Nejjar_2023}. A large-scale survey of over 800 researchers similarly finds that LLMs are often employed for literature-related tasks such as information seeking and editing, but much less so for data cleaning and analysis \cite{Liao_2024}. Finally, Luo et al. (2025) present a broad survey of LLMs for scientific research (LLM4SR), situating such usage within the full research cycle, from hypothesis discovery and experiment planning to writing and peer review \cite{Luo_2025}. Overall, these works suggest that while LLMs hold promise for accelerating and scaling aspects of scientific work, meaningful integration requires strong human oversight and safeguards to maintain scientific standards \cite{Dis_2023}.

There were also some recent impressive achievements for AI-supported scientific research in biology. One example~\cite{Guan_2025} has been described by a liver-disease researcher at Standford Medicine. In that work, AI was asked to scan already available drugs that could potentially treat liver fibrosis. The researcher tasked the AI to search for medicines that are able to influence epigenetic regulators, i.e. that are able to switch  on or off certain genes without changing the DNA itself. The AI mined the biomedical literature and was able to find three promising suggestions. The researcher added two candidates of his own and tested all five drugs in a set of tests on lab-grown liver tissue. Two of drugs suggested by the AI reduced fibrosis and even showed some signs of liver regeneration. None of the researchers suggestions worked. In another example~\cite{Penades_2025}, microbiologists at the Imperial Collage London had an AI agent solve a difficult question about bacterial evolution. The researcher had shown previously that parasitic scraps of DNA could spread within a bacterial population by sticking to and traveling with viruses that infect those bacteria. But some parasitic scraps also appeared in completely different types of bacteria. They asked the AI agent to come up with solutions to this riddle. The agent was given their past research data and related papers. The agent came up with several solutions with one being that DNA fragments might stick also to viruses infecting neighboring bacteria. Recent unpublished data from the researchers are hinting at exactly this mechanism. So the AI came up with a solution in a very short time that took the researchers two years of research.

While the previous examples and related works highlight the potential of AI, there is little work about using AI as a virtual co-scientist for prototyping data analysis and designing intelligent systems. Typically investigations and research about AI for code generation are concentrating only on the pure writing of code. Investigations of AI as scientific agent are mostly concentrating on supporting theoretical findings. The key difference of this case study is that here code generation was performed hand in hand with algorithmic and theoretical problem analysis. This means the AI did not only had to write software or only reason about theoretical frameworks but suggest algorithmic approaches and discuss theoretical background to find a reasonable approach in tandem with the researcher which is then instantiated by programming.

\section{The ELOPE Competition}

The target of the ELOPE (Event-based Lunar OPtical flow Egomotion estimation) competition is to estimate the 3D velocity of a lunar lander from the data of an event camera, an IMU and radar range measurements. Event cameras~\cite{Gallego_2022} are a special type of camera that do not output image frames but pixel events based on brightness changes. Whenever the brightness-change at a pixel exceeds a threshold an event $(x,y,p)$ is generated, where $(x,y)$ is the pixel position of the change and $p$ is the polarity, i.e. telling whether the change was positive or negative. Event cameras promise higher temporal resolution and dynamic range than conventional cameras, sacrificing spatial resolution.

The data set of the ELOPE challenge encompasses 93 simulated landing sequences. For each landing sequence, three major data elements are provided: camera event stream, trajectory and range meter reading.

{\bf Event stream} is a sequence of $(x,y,p,t)$ vectors, where $(x,y)$ is the pixel position, $p$ the polarity and $t$ the timestamp of the event.

{\bf Trajectory} is list of $(x,y,z,v_x,v_y,v_z,\phi,\theta,\psi,p,q,r,t)$ state vectors, where $(x,y,z)$ is the 3D position of the lander in world-coordinates, $(v_x,v_y,v_z)$ is the 3D velocity of the lander, $(\phi,\theta,\psi)$ is the lander's world orientation in Euler angles, $(p,q,r)$ are the angular velocities of the lander and $t$ is the time stamp of the state vector.

{\bf Range meter reading} is a list of distance measures $d$ of a radar attached to the bottom of the lander. During descent of the lunar lander this radar measures the metric distance of the lander to the ground (moon surface).

The sequences are divided into 28 train sequences and 65 test sequences. For train sequences the full data as described above is provided. For test sequences positions $(x,y,z)$ and velocities $(v_x,v_y,v_z)$ are set to \textit{nan}. The goal of the competition was to estimate the missing velocities $(v_x,v_y,v_z)$ for each test trajectory.

\section{Developing an Approach with AI-support}

The competition ran from May to August 2025. We started in the last three weeks of the competition and our team consisted only of one member. To cope with this disadvantage, we decided to engage in a pair-work with ChatGPT (GPT 4.5) for the code generation and the algorithmic design of the approach. This activity yielded some very interesting and some unexpected insight, we'd like to share here with the scientific community.

In this case study, we started completely from scratch. We did not work on this competition, before starting to work on it together with ChatGPT. After some initial reading of the competition explanations, we decided to follow a classical computer vision approach because we had some prior experience with ego-motion estimation. Throughout the development, we mainly used one chat. However, we quickly realized that discussions about alternative approaches poison the further code generation of and discussions with ChatGPT. Also the chat became confusing when discussing several ideas. Hence, we soon began to discuss alternative ideas and solutions in a separate chats. Please note that newer versions of ChatGPT now offer to branch off a discussion, which was not available at the time of the competition (20.05.2025 - 31.08.2025). This new feature certainly improves prototyping as it allows for an easy way to branch off and cut ideas and discussions when it becomes evident they won't work out, without spamming the context window of the chat.

\begin{figure}[t]
  \centering
  \includegraphics[width=\linewidth]{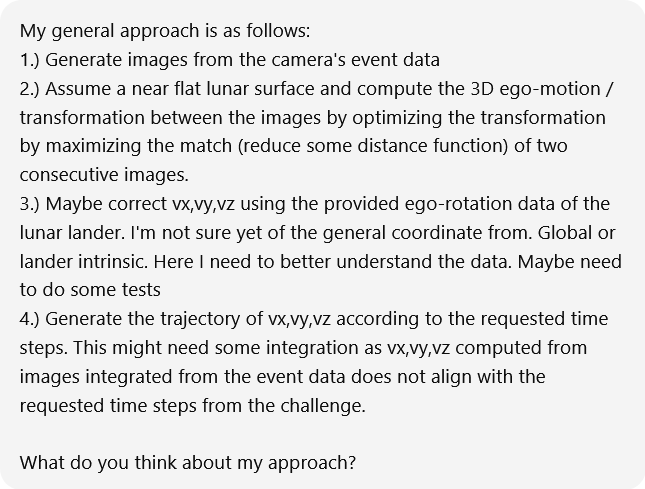}
  \caption{Start of co-development of algorithm for ELOPE challenge.}
  \label{fig:chat-init}
\end{figure}

\begin{figure}[t]
  \centering
  \includegraphics[width=\linewidth]{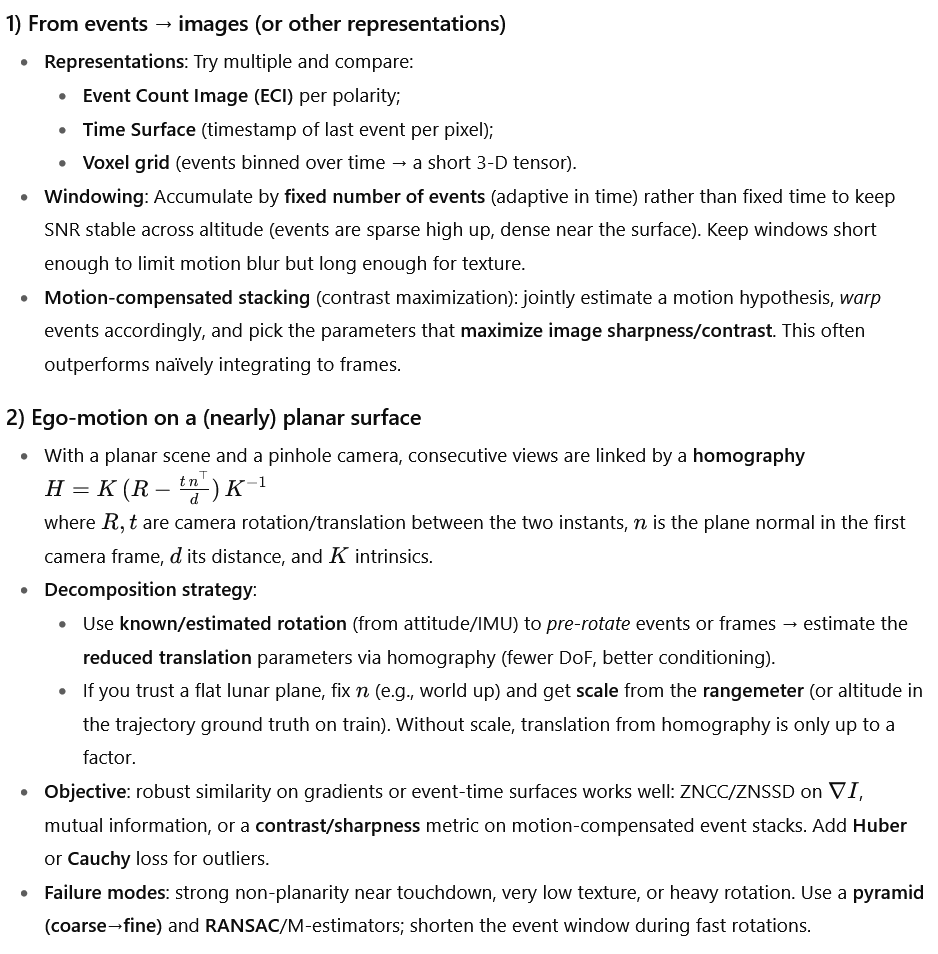}
  \caption{Comments of ChatGPT about proposed approach (see Fig.\ref{fig:chat-init}).}
  \label{fig:comments_approach}
\end{figure}

We started the interaction with giving the chatbot the competition website and telling it we would like to participate. It quickly summarized the competition but with less details about what the actually challenge is. Asking back, it responded with a good summary of the data given, the expected output and the score function. Then we started the actual development by presenting our general idea as shown in Fig.~\ref{fig:chat-init}.
The bot answered with a detailed analysis of the approach with some very important suggestions (see details in Fig.\ref{fig:comments_approach}). For example, for step 1, the integration of events into camera images, it proposed windowing with a fixed number of events instead of time window and to maximize the image contrast. We ignored this proposal for a simpler algorithm and used a fixed time window. However, this contrast maximization was a distinguishing feature of the winning team of the ELOPE challenge. This shows that the LLM was able to link domain specific best practices to the general problem of the competition. Had we followed that advise, we may have catched up with the score of the winning team.

ChatGPT also proposed to compensate the events or image frames with IMU data to improve the stability of the homography estimation. This is also a valuable suggestion that improves performance. To cope with strong rotations and low texture it proposed to apply a pyramid approach. Also a reasonable extension. Finally, it also proposed to separate the polarity. A feature we also had in mind but did not communicate.

In summary, the first analysis of the LLM was surprisingly helpful. We did not follow all the suggestions because we favored to implement a simpler approach given the short remaining time of the competition. However, hindsight it turned out that some of these suggestions may would have led to a score similar to the winning team of the ELOPE competition, highlighting the capability of chatbots to enhance ideation in scientific prototyping.

To keep track of the changes of the LLM we decided to use Git~\cite{Torvalds_2005} as a version control system. Here we used just a local repository. After this initial setup, we told ChatGPT that we now want to start coding the approach. This led ChatGPT to generate the full project at once. However, this did not work out of the box and was hard to debug because of the sheer number of new lines of code. We also had the impression that the code was a very bloated and overly complex. Hence, we decided to restart by asking ChatGPT for a minimal python project setup with \texttt{poetry}. To our surprise this minimal example failed with error although it consisted only of a simple file with a skeleton for a command line interface. ChatPGT also failed to resolve the issue by itself even after several attempts. Asking it for a complete new implementation of the minimal setup directly worked. This is an example of the variable nature of LLMs and necessity to understand and review chatbot output.

With the basic software project structure, we started to implement the approach outlined above step-by-step. The first major implementation was to ask for some code to read the input data, in particular to be able to read in one \texttt{train.npz} file. Here ChatGPT really shined. It generated code that worked out of the box, except for some wrong name of the range data in the \texttt{npz} files. ChatGPT had also added a useful commandline option \code{summarize} which computes some statistics about the data, e.g. the number of events and the average velocity. We did not explicitly asked for this option but found it useful for starting to work with the data set. After giving it the output of the first version of the script, it identified the naming error without being asked explicitly for it fixed it, however, it also added options for reading only a certain number of events or writing the events back out as png. We asked to remove these features, as they did not seem helpful to us. These examples highlight that LLMs need careful supervision, in order to stay focused on the task and prevent them to go astray. Furthermore, it shows that one has to carefully think about the prompt and the context information. If a request is not specific enough (like "{\it generate me some skeleton to start with}") it tends to add unnecessary filler code. Sometimes this is a useful extension, but often this just makes the code unnecessary complex and lengthy. Thus, it is important to monitor the code generation step-by-step and constantly give feedback to the LLM not only about runtime errors to fix but also about bloated code to make it lean. Otherwise, one is quickly lost in sea of generated code.

\begin{figure}[t]
  \centering
  \includegraphics[width=\linewidth]{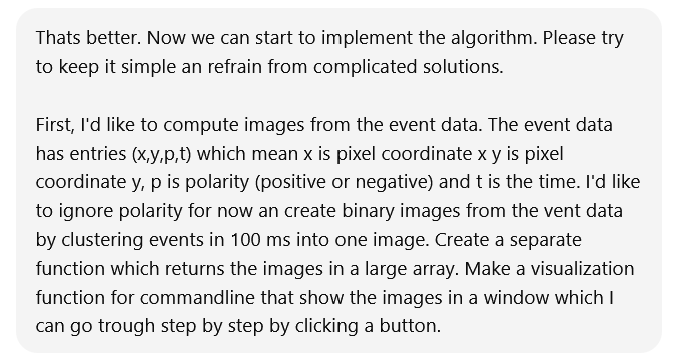}
  \caption{Prompt for asking to implement event agglomeration into images.}
  \label{fig:chat-event-to-image}
\end{figure}

Next we asked for a function to aggregate event camera events into binary images and to provide a visualization of the generated images. As Fig.~\ref{fig:chat-event-to-image} shows, we tried to be more specific with the request.
ChatPGT generated a brief, easy to understand function and added another commandline option \code{view-events} which allowed to step through the images. While the code was good in principle, it had a few flaws that needed some additional iterations. The events where assumed to be in seconds instead of microseconds and the visualization did not work. We asked to redo the visualization with \texttt{matplotlib} and check for the interactive element. ChatGPT asked whether we would like to have the possibility to go back and forth with the arrow keys. This was a great suggestion, as it made debugging and understanding the data much more insightful. From a algorithm development point of view, the implemented visualization is an important asset that gives a quick check whether the implementation of the event-to-image conversion is doing the right thing and was also very useful to judge the impact of parameter changes later on. For example to see what influence the change of the length of the integration window has.

With a working visualization, we could move on to the major element of the approach, the homography~\cite{Hartley_2003} estimation. In a nutshell, a homography describes the perspective changes of a planar surface between two different camera views. This means, we assumed that the lunar surface seen by the camera is nearly flat. From the homography description, movement parameters can be computed. These are up to scale unless an additional hint for the scale is used. In the ELOPE competition the lunar lander's radar range measurements serve as this hint. Since, we were not sure if ChatGPT would foresee the necessary implications of the approach, we explicitly asked to discuss the approach before starting an implementation (see Fig.~\ref{fig:chat-homopgraphy}):
\begin{figure}[t]
  \centering
  \includegraphics[width=\linewidth]{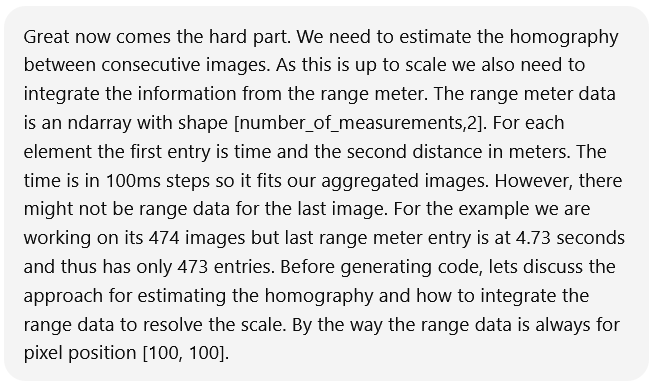}
  \caption{Prompt for asking to discuss the implementation of a homography estimation from the images.}
  \label{fig:chat-homopgraphy}
\end{figure}
While ChatGPT was very proficient in the mathematical foundation of a homography it followed the standard approach of estimating image features for correspondence search, as shown in the response of ChatGPT in Fig.~\ref{fig:chat-homopgraphy-response}.
\begin{figure}[t]
  \centering
  \includegraphics[width=\linewidth]{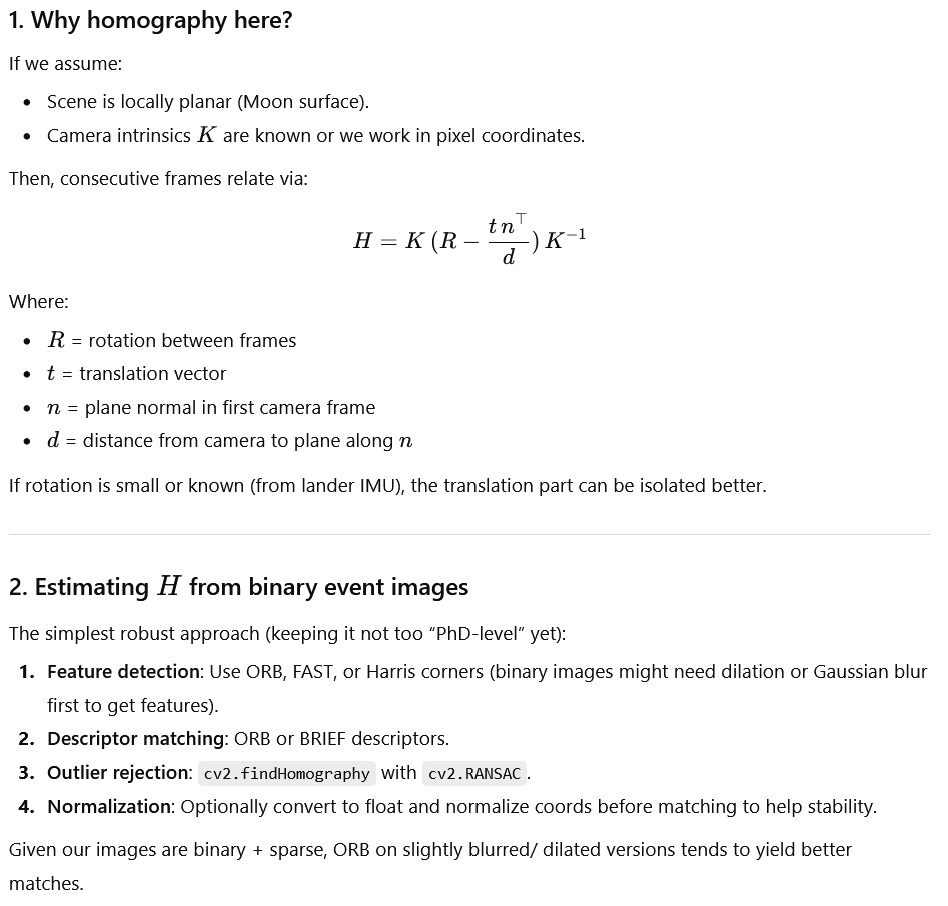}
  \caption{First response to the request for discussing a homography implementation (see Fig.~\ref{fig:chat-homopgraphy}).}
  \label{fig:chat-homopgraphy-response}
\end{figure}
 However, this approach is error prone when working with sparse binary images. Hence, we asked it to reconsider using direct full image fitting. It directly proposed to use an existing implementation of enhanced correlation coefficient (ECC) maximization~\cite{Evangelidis_2008} in \texttt{openCV}~\cite{opencv_library}, an effective method for direct homography estimation from images.

 There was then a lengthy discussion with ChatGPT about how to extract the 3D velocities from the homography. We skip this discussion here for brevity. After considering different options, we came to the conclusion that it is best to estimate the translation vector $\mathbf{t}_t = [\Delta x_t, \Delta y_t, \Delta z_t]^\top$ by applying the homography to the central pixel $(x^c_t, y^c_t)$, as it reveals where this pixel moves to. The difference in positions gives the 2D translations, which are multiplied by the rangefinder measurement to be converted into the 3D translations $\Delta x_t$ and $\Delta y_t$. Similarly, by computing the Jacobian of the center pixel, the scale change $s_t$ can be derived, which corresponds to $\Delta z_t$~\cite{Hartley_2003}. Dividing these three translations by the frame time yields 3D velocities.
 
\begin{figure}[t]
  \centering
  \includegraphics[width=\linewidth]{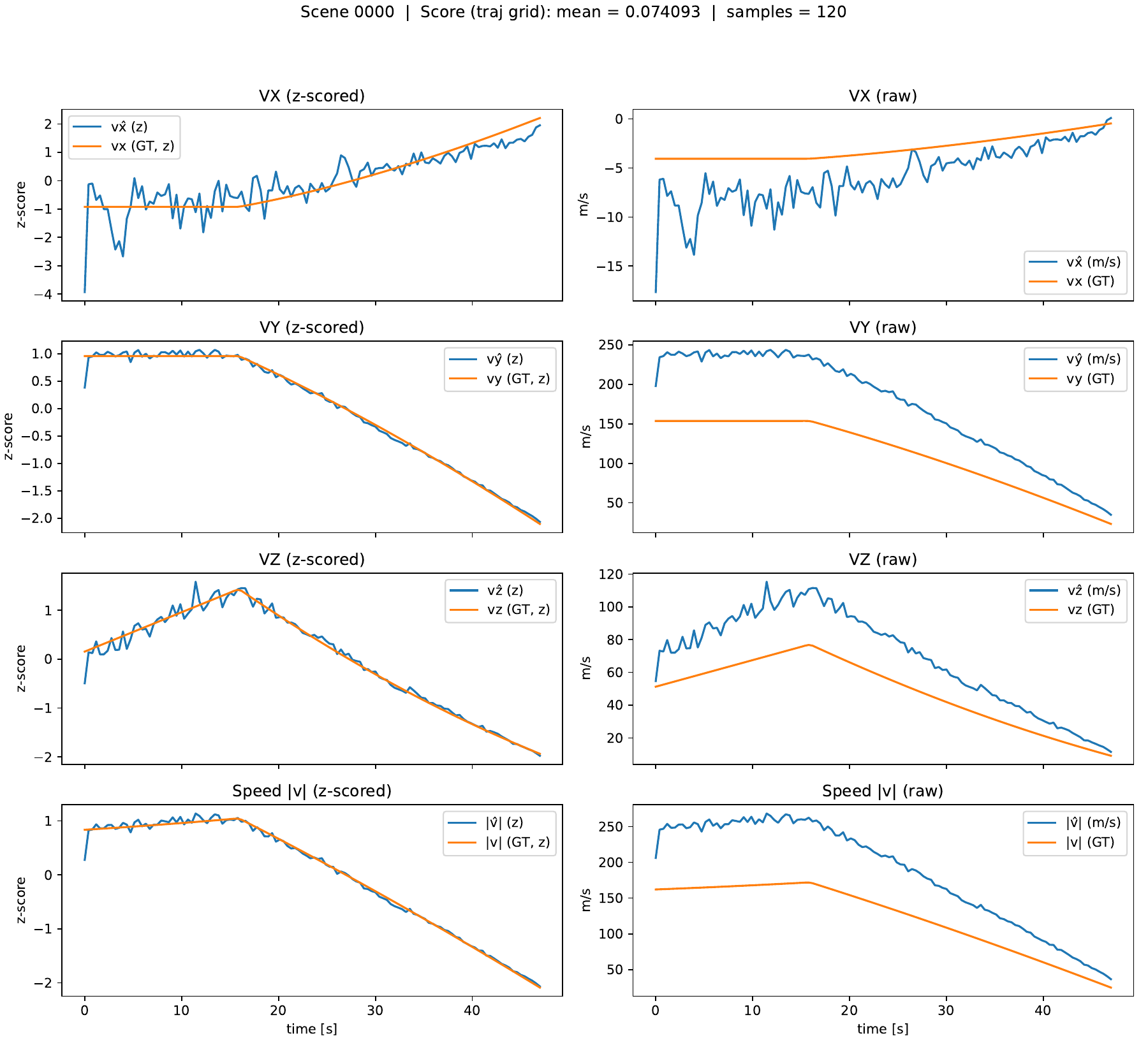}
  \caption{Trajectory visualization with unknown camera matrix. While real velocities show some difference (right column), normalized trajectories overlap nearly perfectly (left column), hinting at a simple additive and multiplicative offset.}
  \label{fig:vis-traj}
\end{figure}

\begin{figure*}[t]
  \centering
  \includegraphics[width=\linewidth]{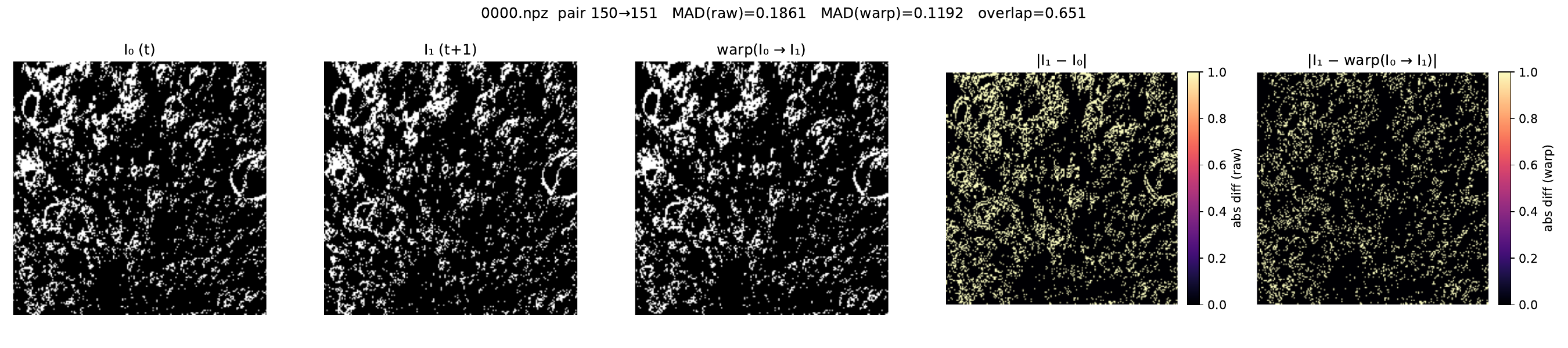}
  \caption{Visualization of raw images and warped images for checking correct working of warping.}
  \label{fig:vis-warp}
\end{figure*}

 One challenge of the competition are the unknown camera parameters. Hence, we asked ChatGPT to continue without camera matrix (i.e. use an identity matrix) and display estimated velocities side-by-side with ground truth of a training stream. The comparison was provided as another commandline option \code{view-compare-vel3d}. With it, we could easily spot sign errors and see if the trajectory is in principle correct, as shown in Fig.\ref{fig:vis-traj}.
 Without asking for it, ChatGPT also computed normalized trajectories and correlations between the computed and real trajectories, which was helping further to spot issues. Additionally, we asked ChatGPT for a visualization of images, images warped by homography and difference images side-by-side to have another visual debugging method (see Fig.~\ref{fig:vis-warp}). Again ChatGPT did a great job in creating this visualization (with another commandline option \code{viz-warp}) but it turned out that it used the homography estimation in the wrong direction (t+1 to t instead of t to t+1), which was obvious do to high dissimilarity between warped and real images. We needed to point it to this error.
 Furthermore, when reviewing parts of the code we realized that images were blurred in a preprocessing step as well as by the ECC function provided by \texttt{openCV}. When mentioning this, ChatGPT set the sigma of the Gaussian blur of the ECC function to 0 instead of removing the preprocessing blurring. This caused the whole estimation to fail as a blurring with sigma 0 leads to blank images. A very hard to debug error if you do not follow code changes. Here we benefited from tracking changes with Git.
 

The last two examples show that one cannot blindly trust the implementations of LLMs unchecked. They have great detailed knowledge of algorithms and underlying theories but seems to lack the ability to understand the correct usage in new applications. Given hints LLMs can easily fix the issue but without explicit feedback they will fail at some point when venturing into uncharted territory. Hence, test-driven implementations are highly recommended, in our view, for current LLMs. They allow for easy verification of LLM code and automatic fixing in agentic settings and enable quick probing of different solutions.

The velocities estimated so far are with respect to the camera but they need to be in world coordinates. Hence, they need to be converted using the known rotation angles. Please note that only a difference in orientation of world and camera coordinate system needs to be considered as the difference in origin is canceled-out for velocities. This compensation was introduced by ChatGPT without errors.
 
\begin{figure}[t]
  \centering
  \includegraphics[width=\linewidth]{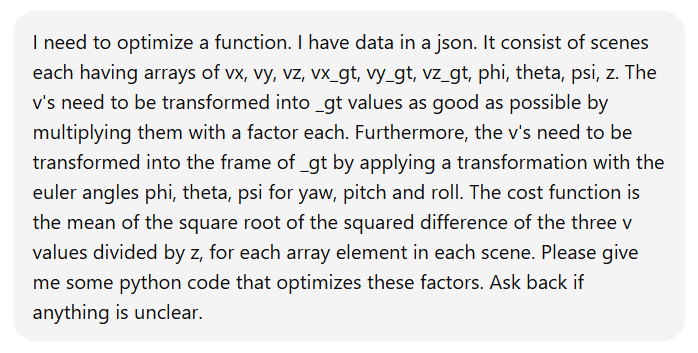}
  \caption{Prompt for asking for code to optimize scale-factors for the velocities.}
  \label{fig:chat-optimize}
\end{figure}

\begin{figure}[t]
  \centering
  \includegraphics[width=\linewidth]{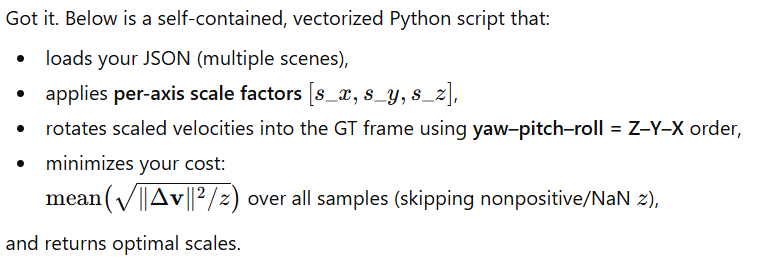}
  \caption{Response to the request for optimizing velocity scale factors.}
  \label{fig:chat-optimize-response}
\end{figure}

After trajectories looked good in general, we had to tackle the unknown camera parameters. We had some issues with finding the camera parameters directly. On the other hand, we had very high correlation values between normalized computed and ground truth trajectories (see Fig.~\ref{fig:vis-traj}). This is a strong hint that values differ only by scale and bias. Bias could be ruled out. Thus, we decided to use optimization to find the scale factor. This task was straightforward and ChatGPT gave optimization code in a few seconds (see prompt and start of response in Fig.~\ref{fig:chat-optimize} and Fig.~\ref{fig:chat-optimize-response}, respectively). Running the optimization took only seconds and yielded factors $[f_x, f_y, f_z] = [0.769, 0.763, 0.832]$.
This demonstrates another great application field for LLMs, optimization. Having a concrete problem to optimize for, LLMs can quickly generate code for optimizing that problem. And since they also have knowledge about the physical principles of the problem, they can suggest which optimization algorithm to choose. 

With the final step of having optimized velocity factors, the algorithm developed in tandem with ChatGPT in about three weeks (one week of work time) resulted in the second best result in the ELOPE challenge. A list of the top five teams score and number of submissions is shown in table~\ref{tab:elope-top5}.
\begin{table}[t]
  \centering
  \caption{Top 5 teams on the ELOPE final results leaderboard, ours being "Team HRI".}
  \label{tab:elope-top5}
  \begin{tabular}{@{}clcc@{}}
    \toprule
    \textbf{Rank} & \textbf{Team} & \textbf{Best Score} & \textbf{\# Submissions} \\
    \midrule
    1 & SOMIS-LAB & 0.006916574694407290 &  9 \\
    2 & Team HRI  & 0.012820211392946732 &  5 \\
    3 & LUNARIS   & 0.015610861273912282 & 15 \\
    4 & caelus    & 0.024244989240780855 & 24 \\
    5 & Apelle    & 0.025399112044236595 &  7 \\
    \bottomrule
  \end{tabular}
\end{table}

\section{Insights}

This section summarizes our insights for using ChatGPT in ESA's ELOPE competition. Due to their similarity, we expect these general observations also to be true for other chatbots and LLMs.

\begin{enumerate}
    \item {\color{ForestGreen} World knowledge makes the chatbot a great co-developer of prototypes. It can assist not only in coding buy also act as a discussion partner for algorithmic theory. Furthermore, it can help in understanding unfamiliar terms when venturing into new fields of research.}
    \item {\color{ForestGreen}The ability of current chatbots to parse and understand website content, shortcuts a lot of explanations. For competitions this also reduces the time to understand details of the competition as the chatbot can quickly parse the text and give a short comprehensive summary. It can also quickly create basic software structure, like data reading or result output based in the website's competition information.}
    \item {\color{ForestGreen}ChatGPT allows for fast implementations of prototypes. It excels at reading of data, writing of results and other basic elements of data engineering.}
    \item {\color{ForestGreen}Finding useful libraries or functions is easier and much faster as compared to classical search engines. One can explain to the chatbot what needs to be done without the necessity to know what libraries are available or how functions are called.}
    \item {\color{ForestGreen}ChatGPT can give useful suggestions and inspirations that help to break habits for coding and developing prototypes. For example it might inspire to use new structures of code, new functions or different kind of visualization.}
    \item {\color{BrickRed} Intermediate discussions poison further output. ChatGPT tends to refer to closed side-track discussion and even puts comments about that in the code.}
    \item {\color{BrickRed} Similar to the tendency of ChatGPT to be too verbose in chat communication, it also tends to create too complex solutions and bloated code.}
    \item {\color{BrickRed} When extending or adapting code, ChatGPT tends to change unrelated things. In most cases these changes are unharmful but increase risk of code instability. For example changes of execution orders, renaming of variables or removal of comments.}
    \item {\color{BrickRed} In longer development cycles, ChatGPT tends to forget important aspects discussed at the beginning. For example it forgot competition constraints defined at the beginning of the chat, so that these constrains had to reiterated again.}
    \item {\color{BrickRed} The internal reasoning of ChatGPT is limited and error prone. Especially, when using seldom function or applying established algorithms to new field, parameters are used or set wrongly. For example it set Gaussian blur to 0, leading to blank images.}
    \item {\color{BrickRed} When discussing unknown problems, ChatGPT tends to make up explanation out of the blue. This can easily lead in the wrong direction during research and development.}
\end{enumerate}

\section{Suggestions}
Based on our experience from the ELOPE competition, we propose the following measures to maximize the success for quick algorithmic and scientific prototype developments when pairing with an LLM:

\subsection{\bf Begin using LLMs for quick prototyping}
The first, suggestion might seem obvious but to our experience a lot of developers are still hesitant to use LLMs or use them just for small code snippets. As discussed in the related work section~\ref{sec:related} LLMs, although being very powerful, have known issues in software development. However, we think their potential with respect to prototyping is overlooked so far. Because they are trained not only on program code but also on massive text corpora on math, physics and computational theories and algorithms, they are able to go beyond mere code generation. In our view they have to potential to be a valuable member in any scientific as well as engineering team.

\subsection{\bf Break out discussions about alternative solutions}
As we mention earlier, we see the major advantage of LLMs in the vast knowledge base they are trained on. Due to this, they not only support in code generation but can also join discussions about algorithmic details and theoretical foundations. In our view, this feature is underrated. However, we found that discussing too much about alternative solutions, poisons the result generations. The LLM tends to take these discussions into account, e.g. by placing comments in the code about alternative solutions. In order to prevent this, we propose to discuss about details and alternatives in a separate chat. New versions of ChatGPT also support branching discussions.

\subsection{\bf Track code changes with a version system}
\label{suggestion-git}
This might seem old-fashioned and unrelated to LLMs, but it is essential to track your code with a versioning system like Git~\cite{Torvalds_2005}. The reason is twofold. First, it is good scientific practice to make your work reproducible. Especially, working on prototypes involves a lot of trail and error of ideas. Being able to jump back to previous stable, working version is essential. Second, as LLMs tend to change more code then necessary, it helps a lot to use a code versioning system for a better visualization of the changes. This way unnecessary or questionable changes can easily be spotted or refined.

\subsection{\bf Use test-driven development}
\label{suggestion-test-driven}
LLMs tend to make very simple but sometimes hard to pinpoint errors. Hence, we recommend to work in a test-driven fashion. Create test defining your software first. This way it is easier to check if the LLM-generated software is doing what it is expected to do and it is easier to check whether new LLM-code-changes break something.

\subsection{\bf Explicitly ask for lean code}
LLMs are known to produce verbose answers~\cite{Saito_2023} and we also found the generated prototype code to be lengthy and complex. In our experience it helped a lot to explicitly ask for lean code. 

\subsection{\bf Go step by step}
It is of course tempting to generate a complete code base all at once. However, it is very hard to understand if you get hundreds of lines of code at once. It might be great if you want to build a web service but it is staggering if you want to develop a prototype where a lot of things are not clear from the beginning. It is much easier to work on and improve a code base whose construction you followed step-by-step. If you don't understand what the code is actually doing you will have hard times to improve it or to find logical errors. This is true in particular when trying to solve a new problem. This, suggestions also goes hand in hand with suggestion \ref{suggestion-git} and \ref{suggestion-test-driven}. Making small changes at a time and committing them in small chunks, allows for easier branching off with new ideas or rolling back to a previous version.

\subsection{\bf Ask for complete functions / files of code}
This is related to the previous suggestions. When changing some smaller things, a chat LLM tends to give you only the changes and you have to insert them yourself at the correct position. This is very error prone and sometimes the LLM has a slightly different code in mind. We found it much easier to ask for updates of complete functions or files and replacing them. Together with change visualization of a versioning system, the changes are much easier to understand and logical or programming errors easier to spot. If you are using an IDE integrated LLM this is not an issue.

\subsection{\bf Raise questions about unclear changes}
Whenever you do not understand a change, ask back. It is well-known that LLMs have a tendency to make things up if they don't know the solution. When you ask back, LLMs are more willing to admit that what they did does not work or was wrong.

\subsection{\bf Ask for visualizations}
Ask the LLM to introduce visualizations at pivotal steps of your algorithm. Understanding the code is important, but understanding the problem and your algorithmic solution is paramount. When developing a prototype for solving a problem, it is in the nature of that process that a lot of things are not clear, e.g. can the method deal with the data noise at hand, is the result accurate enough or was a coordinate conversion performed in the correct way. In many cases such questions are quickly answered by a plot, which helps in getting a deeper understanding of the problem at hand. Before the LLM era such visualization took a decent amount of development time reducing the time for researching ideas, now they are swiftly build and should be used more intensively.

\section{Conclusion}

With the advent of LLMs, workflows in industry and science are about to change. Things that took hours or days in the past, can now be done in minutes. This enables a faster prototyping and testing of ideas in scientific research. However, LLMs are by they nature neither failure-proof nor reproducible. Hence, it is important to take this into account when devising new AI-supported scientific workflows. In this work, we presented our experience in using ChatGPT as a co-scientist when taking part in ESA's ELOPE competition in August 2025. By summarizing the chat and development in chronological order, we showed where ChatGPT excelled and where it failed. Based on our experience we derived a set of measures which we think improve scientific and engineering co-work with LLMs in general. It needs to be noted though that LLM capabilities are currently still subject to large and quick changes and, thus, some of the suggestion might not be applicable in the future. Nevertheless, we hope to provide other researchers and engineers with a good starting point for a successful collaboration with LLMs in their daily work.

\balance

\bibliographystyle{IEEEtran}
\bibliography{bibliography}

@article{Chew_2023,
  title        = {{LLM-Assisted Content Analysis: Using Large Language Models to Support Deductive Coding}},
  author       = {Robert Chew and John Bollenbacher and Michael Wenger and Jessica Speer and Annice Kim},
  journal      = {arXiv preprint arXiv:2306.14924},
  year         = {2023},
  doi          = {10.48550/arXiv.2306.14924},
}

@article{Dis_2023,
  title        = {{ChatGPT: five priorities for research}},
  author       = {Van Dis, Eva AM and Bollen, Johan and Zuidema, Willem and Van Rooij, Robert and Bockting, Claudi L},
  journal      = {Nature},
  volume       = {614},
  number       = {7947},
  pages        = {224--226},
  year         = {2023},
  publisher    = {Nature Publishing Group UK London},
  doi          = {10.1038/d41586-023-00288-7},
}

@online{esa_elope,
  title        = {{ELOPE Challenge --- Kelvins}},
  author       = {{European Space Agency}},
  year         = {2025},
  url          = {https://kelvins.esa.int/elope/},
}

@article{Evangelidis_2008,
  title        = {{Parametric Image Alignment Using Enhanced Correlation Coefficient Maximization}}, 
  author       = {Evangelidis, Georgios D. and Psarakis, Emmanouil Z.},
  journal      = {IEEE Transactions on Pattern Analysis and Machine Intelligence}, 
  volume       = {30},
  number       = {10},
  pages        = {1858-1865},
  year         = {2008},
  doi          = {10.1109/TPAMI.2008.113},
}

@article{Fanti_2026,
  title        = {{Event-based lunar optical flow egomotion estimation challenge: design and results of the ELOPE competition}},
  author       = {Fanti, Pietro and Williams, Leon BS and Dvo{\v{r}}{\'a}k, Ond{\v{r}}ej and M{\"a}rtens, Marcus and Chin, Tat-Jun and Ji, Hongbo and Chen, Bofei and Xie, Dongyu and Qiao, Kaifan and Li, Bohao and Einecke, Nils and Arumugam, Subramanian and Padhy, Amulya Ratna and Rath, Sumeet Kumar and Mahapatra, Swati Sonal and Izzo, Dario},
  journal      = {npj Space Exploration},
  volume       = {2},
  number       = {1},
  pages        = {18},
  year         = {2026},
  doi          = {10.1038/s44453-026-00033-0},
}

@article{Gallego_2022,
  title        = {{Event-Based Vision: A Survey}}, 
  author       = {Gallego, Guillermo and Delbrück, Tobi and Orchard, Garrick and Bartolozzi, Chiara and Taba, Brian and Censi, Andrea and Leutenegger, Stefan and Davison, Andrew J. and Conradt, Jörg and Daniilidis, Kostas and Scaramuzza, Davide},
  journal      = {IEEE Transactions on Pattern Analysis and Machine Intelligence}, 
  year         = {2022},
  volume       = {44},
  number       = {1},
  pages        = {154-180},
  doi          = {10.1109/TPAMI.2020.3008413},
}

@article{Guan_2025,
  title        = {{AI-Assisted Drug Re-Purposing for Human Liver Fibrosis}},
  author       = {Guan, Yuan and Cui, Lu and Inchai, Jakkapong and Fang, Zhuoqing and Law, Jacky and Brito, Alberto Alonzo Garcia and Pawlosky, Annalisa and Gottweis, Juraj and Daryin, Alexander and Myaskovsky, Artiom and Ramakrishnan, Lakshmi and Palepu, Anil and Kulkarni, Kavita and Weng, Wei-Hung and Cheng, Zhuanfen and Natarajan, Vivek and Karthikesalingam, Alan and Rong, Keran and Xu, Yunhan and Tu, Tao and Peltz, Gary},
  journal      = {Advanced Science},
  year         = {2025},
  pages        = {e08751},
  doi          = {10.1002/advs.202508751},
}

@book{Hartley_2003,
  title        = {{Multiple view geometry in computer vision}},
  author       = {Hartley, Richard and Zisserman, Andrew},
  year         = {2003},
  publisher    = {Cambridge university press}
}

@article{Ifargan_2025,
  title         = {Autonomous llm-driven research—from data to human-verifiable research papers},
  author        = {Ifargan, Tal and Hafner, Lukas and Kern, Maor and Alcalay, Ori and Kishony, Roy},  
  journal       = {NEJM AI},
  volume        = {2},
  number        = {1},
  pages         = {AIoa2400555},
  year          = {2025},
  doi           = {10.1056/AIoa2400555},
}

@inproceedings{Kazemitabaar_2024,
  title        = {{How Novices Use LLM-Based Code Generators to Solve CS1 Coding Tasks in a Self-Paced Learning Environment}},
  author       = {Majeed Kazemitabaar and Xinying Hou and Austin Henley and Barbara J. Ericson and David Weintrop and Tovi Grossman},
  booktitle    = {Proceedings of the 23rd Koli Calling International Conference on Computing Education Research},
  year         = {2024},
  doi          = {10.1145/3631802.3631806},
}

@article{Joublin_2023,
  title        = {{A glimpse in ChatGPT capabilities and its impact for AI research}},
  author       = {Joublin, Frank and Ceravola, Antonello and Deigmoeller, Joerg and Gienger, Michael and Franzius, Mathias and Eggert, Julian},
  journal      = {arXiv preprint arXiv:2305.0608},
  year         = {2023},
  doi          = {10.48550/arXiv.2305.06087},
}

@article{Liao_2024,
  title         = {{LLMs as Research Tools: A Large Scale Survey of Researchers’ Usage}},
  author        = {Liao, Zhehui and Antoniak, Maria and Cheong, Inyoung and Cheng, Evie Yu-Yen and Lee, Ai-Heng and Lo, Kyle and Chang, Joseph Chee and Zhang, Amy X},
  journal       = {arXiv preprint arXiv:2411.05025},
  year          = {2024},
  doi           = {10.48550/arXiv.2411.05025},
}

@article{Luo_2025,
  title         = {{LLM4SR: A Survey on Large Language Models for Scientific Research}},
  author        = {Luo, Yuxi and Zhang, Yichong and Wang, Zhi and others},
  journal       = {arXiv preprint arXiv:2501.04306},
  year          = {2025},
  doi           = {10.48550/arXiv.2501.04306},
}

@article{Miserendino_2025,
  title        = {{SWE-Lancer: Can Frontier LLMs Earn \$1 Million from Real-World Freelance Software Engineering?}},
  author       = {Samuel Miserendino and Michele Wang and Tejal Patwardhan and Johannes Heidecke},
  journal      = {arXiv preprint arXiv:2502.12115},
  year         = {2025},
  doi          = {10.48550/arXiv.2502.12115},
}

@inproceedings{Nam_2024,
  title        = {{Using an LLM to Help With Code Understanding}},
  author       = {Daye Nam and Andrew Macvean and Vincent Hellendoorn and Bogdan Vasilescu and Brad Myers},
  booktitle    = {Proceedings of the IEEE/ACM 46th International Conference on Software Engineering},
  year         = {2024},
  pages        = {1--13},
  doi          = {10.1145/3597503.3639187},
}

@inproceedings{Naman_2025,
  title        = {{Analysis of Student-LLM Interaction in a Software Engineering Project}},
  author       = {Agrawal Naman and Ridwan Shariffdeen and Guanlin Wang and Sanka Rasnayaka and Ganesh Neelakanta Iyer},
  booktitle    = {2025 IEEE/ACM International Workshop on Large Language Models for Code (LLM4Code)}, 
  pages        = {112-119},
  year         = {2025},
  doi          = {10.1109/LLM4Code66737.2025.00019},
}

@article{Nejjar_2023,
  title        = {{LLMs for science: Usage for code generation and data analysis}},
  author       = {Nejjar, Mohamed and Zacharias, Luca and Stiehle, Fabian and Weber, Ingo},
  journal      = {Journal of Software: Evolution and Process},
  volume       = {37},
  number       = {1},
  pages        = {e2723},
  year         = {2025},
  doi          = {10.1002/smr.2723},
}

@article{opencv_library,
  title         = {{The OpenCV Library}},  
  author        = {Gary Bradski},
  journal       = {Dr. Dobb's Journal of Software Tools},
  volume        = {25},
  number        = {11},
  pages         = {120, 122--125},
  year          = {2000},
}

@article{Penades_2025,
  title        = {{AI mirrors experimental science to uncover a mechanism of gene transfer crucial to bacterial evolution}},
  author       = {Penad{\'e}s, Jos{\'e} R and Gottweis, Juraj and He, Lingchen and Patkowski, Jonasz B and Daryin, Alexander and Weng, Wei-Hung and Tu, Tao and Palepu, Anil and Myaskovsky, Artiom and Pawlosky, Annalisa and others},
  journal      = {Cell},
  year         = {2025},
  publisher    = {Elsevier},
  doi          = {10.1016/j.cell.2025.08.018},
}

@article{Peng_2023,
  title        = {{The Impact of AI on Developer Productivity: Evidence from GitHub Copilot}},
  author       = {Sida Peng and Eirini Kalliamvakou and Peter Cihon and Mert Demirer},
  journal      = {arXiv preprint arXiv:2302.06590},
  year         = {2023},
  doi          = {10.48550/arXiv.2302.06590},
}

@article{Ren_2025,
  title         = {{Towards Scientific Intelligence: A Survey of LLM-based Scientific Agents}},
  author        = {Ren, Shuo and Jian, Pu and Ren, Zhenjiang and Leng, Chunlin and Xie, Can and Zhang, Jiajun},
  journal       = {arXiv preprint arXiv:2503.24047},
  year          = {2025},
  doi           = {10.48550/arXiv.2503.24047},
}

@inproceedings{Sadik_2025,
  title         = {{Benchmarking LLM for Code Smells Detection: OpenAI GPT-4.0 vs DeepSeek-V3}},
  author        = {Sadik, Ahmed R and Govind, Siddhata},
  booktitle     = {Proceedings of the 29th International Conference on Evaluation and Assessment in Software Engineering},
  pages         = {969--975},
  year          = {2025},
  doi           = {10.1145/3756681.3756993},
}

@article{Saito_2023,
  title         = {{Verbosity Bias in Preference Labeling by Large Language Models}}, 
  author        = {Keita Saito and Akifumi Wachi and Koki Wataoka and Youhei Akimoto},
  journal       = {arXiv preprint arXiv:2310.10076},
  year          = {2023},
  doi           = {10.48550/arXiv.2310.10076},
}

@inproceedings{Schmidgall_2025,
  title         = {{Agent Laboratory: Using LLM Agents as Research Assistants}},
  author        = {Schmidgall, Samuel and Su, Yusheng and Wang, Ze and Sun, Ximeng and Wu, Jialian and Yu, Xiaodong and Liu, Jiang and Zicheng Liu and Barsoum, Emad},
  booktitle     = {arXiv preprint arXiv:2501.04227},
  year          = {2025},
  doi           = {10.48550/arXiv.2501.04227},
}

@misc{Torvalds_2005,
  title        = {{Git -- Fast Version Control System}},
  author       = {Linus Torvalds and Junio C. Hamano and {Git Contributors}},
  howpublished = {\url{https://git-scm.com/}},
  year         = {2005}
}

@inproceedings{Zi_2025,
  title        = {{“I Would Have Written My Code Differently”: Beginners Struggle to Understand LLM-Generated Code}},
  author       = {Yangtian Zi and Luisa Li and Arjun Guha and Carolyn Anderson and Molly Q. Feldman},
  booktitle    = {Proceedings of the 33rd ACM International Conference on the Foundations of Software Engineering},
  pages        = {1479–-1488},
  year         = {2025},
  doi          = {10.1145/3696630.3731663},
}

\end{document}